\documentclass[10pt,twocolumn,letterpaper]{article}

\usepackage{cvpr}
\usepackage{times}
\usepackage{epsfig}
\usepackage{graphicx}
\usepackage{amsmath}
\usepackage{amssymb}
\usepackage[dvipsnames]{xcolor}
\usepackage{booktabs}
\usepackage{subfig}
\usepackage{mathtools}
\usepackage{authblk}

\usepackage[pagebackref=true,breaklinks=true,letterpaper=true,colorlinks,bookmarks=false]{hyperref}

\def\model{Discriminative Relational Recurrent Network\xspace}
\def\modelshort{DR$^2$N\xspace}

\cvprfinalcopy 

\def\cvprPaperID{1745} 

\ifcvprfinal\fi
\begin{document}

\title{Relational Action Forecasting}

\author[1]{Chen Sun}
\author[2]{Abhinav Shrivastava}
\author[1]{Carl Vondrick}
\author[1]{\\Rahul Sukthankar}
\author[1]{Kevin Murphy}
\author[1]{Cordelia Schmid}
\affil[1]{Google Research}
\affil[2]{University of Maryland}
\maketitle

\begin{abstract}
This paper focuses on multi-person action forecasting in videos. More precisely, given a history of $H$ previous frames, the goal is to detect actors and to predict their future actions for the next $T$ frames. Our approach jointly models temporal and spatial interactions among different actors by constructing a recurrent graph, using actor proposals obtained with Faster R-CNN as nodes. Our method learns to select a subset of discriminative relations without requiring explicit supervision, thus enabling us  to tackle challenging visual data. We refer to our model as \model (\modelshort). Evaluation of action prediction on AVA  demonstrates the effectiveness of our proposed method compared to simpler baselines. Furthermore, we significantly improve performance on the task of early action classification on J-HMDB, from the previous SOTA of 48\% to 60\%.
\end{abstract}

\section{Introduction}

\begin{figure}[t]
    \centering
    \includegraphics[width=.95\linewidth]{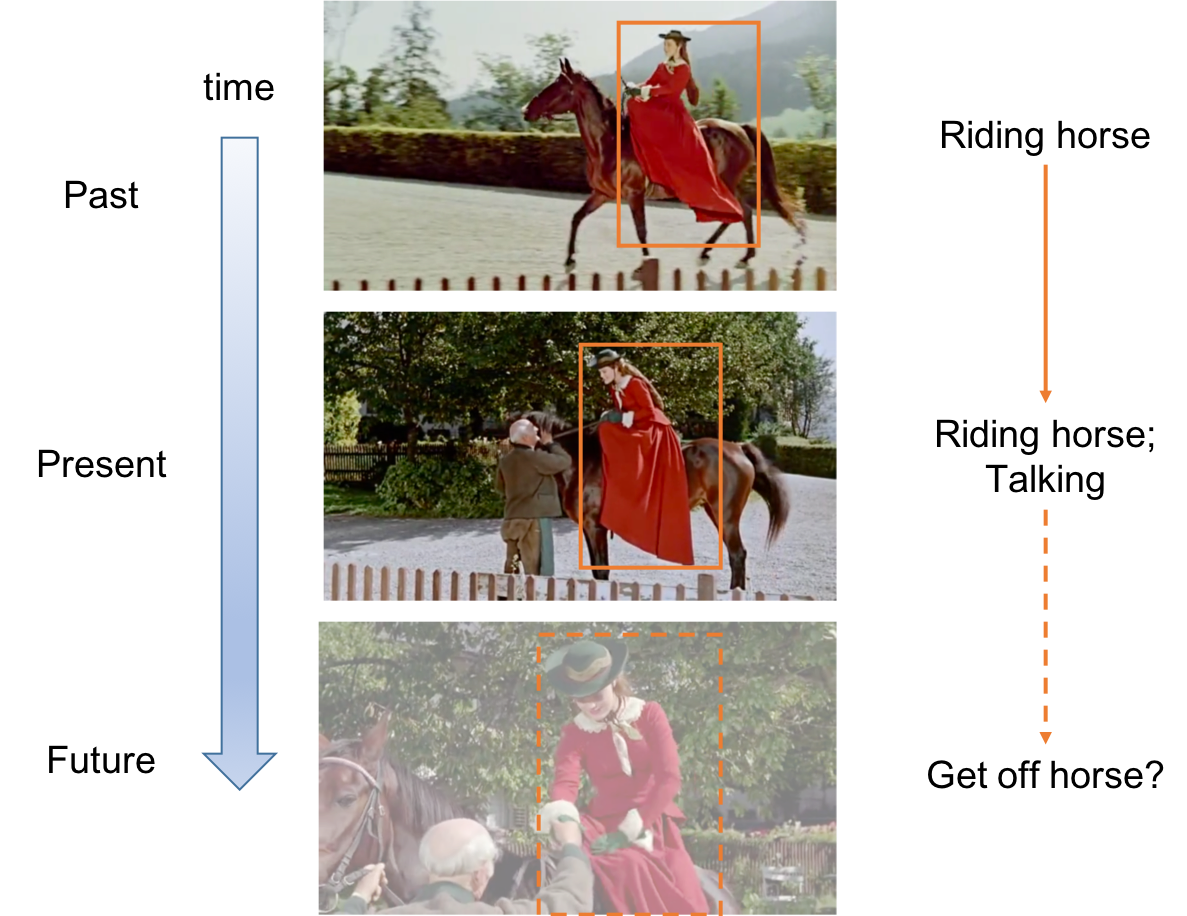}
    \caption{Action prediction from a single frame (middle) is ambiguous, but requires temporal context for the actors and their interactions. It only becomes apparent that the lady will get off the horse, if we know that she was riding towards the man, and the man is holding the horse.}
    \label{fig:teaser}
\end{figure}

In this paper, we consider the task of forecasting what high level actions people will perform in the future, given noisy visual evidence.
For example, consider Figure~\ref{fig:teaser}: given
the current video frame,
and a sequence of past frames, 
we would like to  detect (localize) the people (man and woman), and classify their current actions (woman rides horse, man and woman are talking), as well as predict future plausible action sequences for each person (woman will get off horse, man will hold horse reigns).

More formally, 
our task is to compute the probability $p(N, b^0_{1:N}, a_{1:N}^{0:T} | V^{-H:0})$,
where $V^t$ is the frame at time $t$ (where $t=0$ is the present),
$V=V^{-H:0}$ is the visual history of $H$ previous frames,
$N$ is the number of predicted actors,
$b_n^0$ is the predicted location (bounding box) of actor $n$ at time 0,
and $a_n^t$ is the predicted action label for actor $n$ at time $t$,
which we compute for $t=0:T$, where $T$ is the maximum forecasting horizon.
This formulation is closely related to but different than prior work.
In particular, 
video classification focuses on computing a single global label in the offline scenario,
$p(c|V^{0:T})$;
spatio-temporal action detection focuses on  multi-agent localization and classification,
but in the offline scenario, $p(a_{1:N}^{0:T}, b_{1:N}^{0:T}|V^{0:T})$;
action anticipation focuses on the future, but for a single class label, $p(c^{0:T}|V^{-H:0})$;
and
trajectory forecasting focuses on multiple agents in the future, but generally 
focuses on locations, not actions, and assumes that the past locations (and hence the number of agents) is observed:
$p(b_{1:N}^{1:T}|V^{-H:0}, b_{1:N}^{-H:0}, N)$.
By contrast, we only observe past frames, and want to predict future high level actions of agents.
This could be useful for self-driving cars, human-robot interaction, etc.
See Section~\ref{sec:related} for a more detailed discussion of related work.

\begin{figure*}[ht]
    \centering
    \includegraphics[width=1.\linewidth]{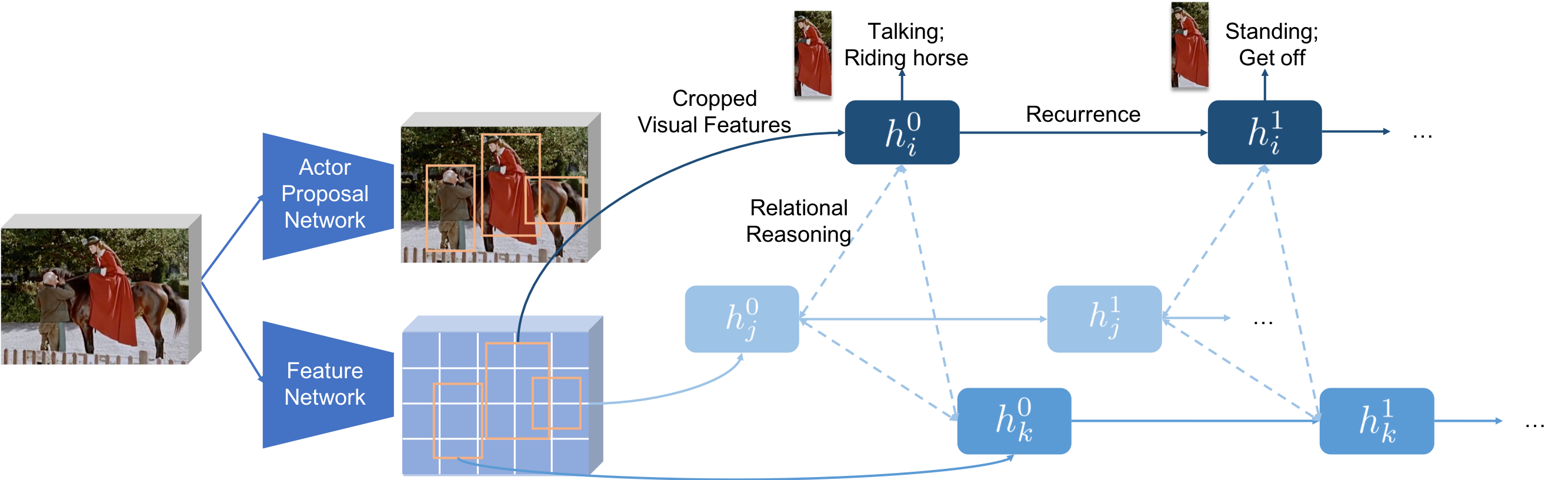}
    \caption{Overview of our approach \model (\modelshort). Given actor proposals and their spatio-temporal descriptors at a given time T=0, we model their relations by a graph neural network and its recurrence over time (here with a GRU).}
    \label{fig:flow_chart}
\end{figure*}

Our proposed approach is to create a graph-structured recurrent neural network (GRNN), in which nodes correspond to candidate person detections (from an object detector trained on person examples), and edges represent potential interactions between the people.
Each node has an action label and bounding box location associated with it. 
(Note that some nodes may be false positive detections arising from the detector,
and should be labeled as such.)
We use a modified version of graph attention networks \cite{Velickovic2018}
applied to a fully connected pairwise graph
to capture interaction effects.
Nodes are also linked over time via RNNs
to capture temporal patterns in the data.
 We name our proposed framework \model (\modelshort).
 See Figure~\ref{fig:flow_chart} for an illustration. 

We train the model in  a weakly supervised way, in the sense that we have a ground truth set of labeled bounding boxes for
people in frame 0,
and we have action labels (but not location labels)
for the people for frames $1:T$. However, we do not observe any edge information. 
Our model is able to learn which edges are useful, so as to maximize node classification performance.
On the challenging AVA dataset~\cite{ava_cvpr18},
we show that our model outperforms various strong baselines at the task of predicting
person action labels for up to 5 seconds into the future.

To be more directly comparable to prior work,
we also consider another task,
namely ``early classification'' of video clips,
where the task is to compute $p(c | V^{0:t})$, where $c$ is the class label for the video clip of length $T$,
and $t < T$ is some prefix of the clip (we use the first 10\% to 50\% of the frames).
We modify our model for this task, and train it on the J-HMDB dataset,
as used in prior works \cite{Singh_ICCV2017,Soomro_online}.
We achieve significant improvements over the previous state of the art,
increasing the classification accuracy given a 10\% prefix
from 48\% (obtained by \cite{Singh_ICCV2017}) to 60\%.

\section{Related work}
\label{sec:related}

\noindent\textbf{Action recognition.} Human action recognition in videos is dominated primarily by three well-established tasks: action classification~\cite{schuldt2004recognizing, activitynet, Marszalek2009, jhmdb, ucf101, youtube8m, Karpathy2014, monfort2018moments, kinetics17}, temporal action localization~\cite{activitynet, THUMOS,MultiTHUMOS,charades2016}, and spatio-temporal action detection~\cite{Ke2005,Yuan2009,ucfsports,jhmdb,ucf101,weinzaepfel2015,mettes2016,ava_cvpr18}. 
Given a video (a set of frames), the goal of action classification is to assign action labels to the entire video, whereas temporal action localization assigns labels to only a subset of frames representing the action. Spatio-temporal action detection combines the temporal action localization with actor detection, i.e., detecting actors per-frame in a subset of frames and assigning action labels to per-actor spatio-temporal tubes~\cite{Ke2005,Yuan2009,jhmdb}. 

The three standard tasks discussed above assume that the entire video is observed, therefore prediction in any frame can utilize past or future frames. In contrast, this paper introduces the task of actor-level action prediction, i.e., given a video \textit{predict} or \textit{anticipate} what actions \textit{each actor} will perform in the future.
This task operates in a more practical setup where only the past and present frames are observed, and predictions have to be made for unobserved future frames. Note that the proposed task inherently requires spatio-temporal actor detection in the observed frames. 

\noindent\textbf{Future prediction.} Our work follows a large number of works studying future prediction~\cite{VondrickNIPS16,petrovic2006recursive,mathieu2015deep,ranzato2014video,finn2016unsupervised,babaeizadeh2017stochastic,kalchbrenner2016video,lotter2016deep,xue2016visual,walker2016uncertain,srivastava2015unsupervised,walker2014patch,yuen2010data,vondrick2016anticipating,kitani2012activity,fragkiadaki2015recurrent,zhou2015temporal,alahi2016social,robicquet2016learning,walker2015dense,villegas2017learning,fragkiadaki2017motion,DESIRE, EPICKITCHENS, MultiTHUMOS, luc2017predicting}. Broadly speaking, the research on future prediction follows two main themes: generating future frame(s)~\cite{VondrickNIPS16,petrovic2006recursive,mathieu2015deep,ranzato2014video,finn2016unsupervised,babaeizadeh2017stochastic,kalchbrenner2016video,lotter2016deep,xue2016visual,srivastava2015unsupervised} and predicting future labels or state(s)~\cite{walker2014patch,yuen2010data,vondrick2016anticipating,kitani2012activity,fragkiadaki2015recurrent,zhou2015temporal,alahi2016social,robicquet2016learning,villegas2017learning,fragkiadaki2017motion,DESIRE, EPICKITCHENS, MultiTHUMOS, luc2017predicting,walker2016uncertain,CarNet18}. For future frame generation, there is a wide variety of approaches ranging from predicting intermediate representations (e.g., optical flow~\cite{ranzato2014video,fragkiadaki2017motion,walker2015dense}, human pose~\cite{villegas2017learning,walker2017pose}) and using it to generate future pixels, to directly generating future pixels by extending generative models for images to videos~\cite{VondrickNIPS16,petrovic2006recursive,finn2016unsupervised,babaeizadeh2017stochastic}. Though the quality of generated frames has improved over the past few years~\cite{vondrick2017generating,babaeizadeh2017stochastic}, 
it is arguably solving a much harder task than necessary (after all, humans can predict likely future actions, but cannot predict likely future pixels).

The second theme in future prediction circumvents pixel generation and directly predicts future states~\cite{walker2014patch,yuen2010data,vondrick2016anticipating,kitani2012activity,fragkiadaki2015recurrent,zhou2015temporal,alahi2016social,robicquet2016learning,villegas2017learning,fragkiadaki2017motion,DESIRE, EPICKITCHENS, MultiTHUMOS, luc2017predicting,walker2016uncertain,CarNet18}. These future states can vary from low-level trajectories~\cite{walker2014patch,kitani2012activity,alahi2016social,robicquet2016learning,CarNet18,DESIRE} for different agents to high-level semantic outputs~\cite{luc2017predicting,EPICKITCHENS,MultiTHUMOS,ma2016learning,su2017predicting}. The trajectory-based approaches rely on modeling and predicting agent behavior in the future, where an agent can be an object (such as human~\cite{kitani2012activity,alahi2016social,robicquet2016learning} or car~\cite{CarNet18}) or an image patch~\cite{walker2014patch}. Most of these methods require a key assumption that the scene is static; i.e., no camera motion (e.g., VIRAT dataset~\cite{oh2011large}) and no movement of non-agent entities~\cite{kitani2012activity,alahi2016social,robicquet2016learning,CarNet18}. Our method is similar to these trajectory-based methods, in the sense that our state prediction is also about agents. However, we do not  assume anything about the scene.
In particular, the AVA dataset has significant camera motion and scene cuts, in addition to agent motion and cluttered, dynamic backgrounds.

Much work on future forecasting focuses on the spatop-temporal extent of detected actions~\cite{Singh_ICCV2017,Soomro_online,hoai2014max,ma2016learning}.
However, there is some work on forecasting  high-level semantic states,
ranging from semantic segmentation masks~\cite{luc2017predicting} to action classes~\cite{EPICKITCHENS,MultiTHUMOS}. 
However, our method has several key distinctions. First, many works require the semantic states 
to be part of the input to the forecasting algorithm,  whereas our method detects actors and actions from past/present frames.
Second, most of the state prediction methods operate on MoCap data (e.g., 3D human keypoints)~\cite{fragkiadaki2015recurrent,jain2016structural},
whereas our approach works from  pixels. 
Third, many method assume static cameras and possibly single agents,
whereas we can forecasting labels for multiple agents in unconstrained video.

\noindent\textbf{Relational reasoning.} Our proposed approach  builds on the field of relational reasoning~\cite{RN_survey,RN_deepmind17,RRN,kipf2018neural,Santoro2018RelationalRN,Velickovic2018,rn_od,nonlocal_cnn,dai_cvpr17,Hamilton2017,battaglia2016interaction,ORN,RRN}. This is natural because the current and future actions of an actor rely heavily on the dynamics it shares with other actors~\cite{alahi2016social}. Relational reasoning~\cite{RN_survey}, in general, can capture relationships between a wide array of entities. For example, relationship between abstract entities or features~\cite{RN_deepmind17,nonlocal_cnn}, different objects~\cite{ORN}, humans and objects~\cite{hico2015,hico_det_wacv2018,gkioxari2017interactnet,yao2010modeling}, humans and context~\cite{acrn}, humans and humans~\cite{alahi2016social,ibrahim2018hierarchical}, etc. Our work aims to capture human-human relationships to reason about future actions. 

In terms of modeling these relationships, the standard tools include Interaction Network (IN)~\cite{battaglia2016interaction}, Relation Network (RN)~\cite{RN_deepmind17},  Graph Neural Network (GNN)~\cite{Hamilton2017}, 
Graph ATtenion networks \cite{Velickovic2018},
as well as their contemporary extensions to videos, such as Actor-centric Relation Network (ACRN)~\cite{acrn} and Object Relation Network (ORN)~\cite{ORN}. Similar to ACRN and ORN, our proposed DR$^2$N tries to capture relation between different entity in videos for spatio-temporal reasoning. However, as opposed to modeling exhaustive relationships (RN~\cite{RN_deepmind17} and ORN~\cite{ORN}), our method discovers discriminative relationships for the task of actor-level action prediction. Compared to ACRN, our method  focuses on forecasting future actor labels given past visual information. In addition, we model interaction between the agents, but ignore any objects in the scene.

\section{Approach}

In this section, we describe our approach in more detail.

\newcommand{\fRPN}{f_{\mathrm{RPN}}}
\newcommand{\fGNN}{f_{\mathrm{GNN}}}
\newcommand{\fRNN}{f_{\mathrm{RNN}}}
\newcommand{\fROI}{f_{\mathrm{ROI}}}
\newcommand{\fCLS}{f_{\mathrm{CLS}}}
\newcommand{\fVIDEO}{f_{\mathrm{S3D}}}
\newcommand{\fnode}{f_{\mathrm{node}}}
\newcommand{\fedge}{f_{\mathrm{edge}}}
\newcommand{\fattn}{f_{\mathrm{attn}}}
\newcommand{\Cat}{\mathrm{Cat}}
\newcommand{\softmax}{\mathcal{S}}
\newcommand{\be}{\begin{equation*}}
\newcommand{\ee}{\end{equation*}}
\newcommand{\defeq}{\triangleq}

\subsection{Creating the nodes in the graph}

 To generate actor proposals and extract the initial actor-level features for action prediction, we build our system on top of the two-stage Faster RCNN~\cite{ren2015faster} detector. The first stage is a region proposal network (RPN) that generates a large set (\eg hundreds) of candidate actor proposals in terms of 2D bounding boxes on a single frame. We apply this RPN module on the last observed frame $V^0$
 to locate actors whose actions are to be predicted. These become nodes in the graph.
 
 The second stage associates features with these nodes.
 We first extract visual features from video inputs, $V^{-H:0}$, using a 3D CNN (see \ref{sec:details} for details), and 
 then crop out features inside the bounding boxes for all (candidate) actors using ROIPooling. Let $\mathbf{v}_i$ be the visual features
for actor $i$.

The third stage is to connect the nodes together into a fully connected graph.
However, since not all nodes are equally useful for prediction, in Section~\ref{sec:drn} we explain how to learn discriminative edge weights.

\subsection{Modeling node dynamics}
\label{sec3:rnn}

We model the action dynamics of individual actors using RNNs. Given $\mathbf{h}_i^t$ as the latent representation for actor $i$ at time $t$ and $\mathbf{a}_i^t$ as the set of actor labels, we have
\begin{equation}
\mathbf{h}_i^t = \fRNN(\mathbf{h}_i^{t-1}, \mathbf{a}_i^{t-1})
\end{equation}
\begin{equation}
\mathbf{a}_i^t = \fCLS(\mathbf{h}_i^{t})
\end{equation}
where  $\fRNN(\cdot)$ is the RNN update function, and $\fCLS(\cdot)$ is an action classifier that decodes the latent states $\mathbf{h}$ into action labels (we use a simple MLP for this, see \ref{sec:details} for details).
The initial state $\mathbf{h}_i^0$ is set to $\mathbf{v}_i$, the visual features extracted for this bounding box.

To make the model scalable over a varying number of actors, the RNN function $\fRNN(\cdot)$ is shared over all actors. Similarly, the action classifier $\fCLS(\cdot)$ is shared over all actors and all time steps.

\begin{table*}
\begin{align*}
    p(N, b^0_{1:N}, a_{1:N}^{0:T} | V^{-H:0})
    &= \delta(N, b_{1:N}^0| \fRPN(V^0))
    p(a_{1:N}^{0}, h_{1:N}^0 | b_{1:N}^0, V^{-H:0}) 
    \prod_{t=1}^T 
p(a_{1:N}^{t}, h_{1:N}^t|a_{1:N}^{1:t-1}, h_{1:N}^{t-1}) 
\\
p(a_{1:N}^{0}, h_{1:N}^0|b_{1:N}^0,V)
 &= \prod_{n=1}^N \Cat(a_n^0|\fCLS(h_n^0))
 \delta(h_n^0 | \fROI(\fVIDEO(V^{-H:0}), b_n^0)) 
 \\
  p(a_{1:N}^{t}, h_{1:N}^t|a_{1:N}^{t-1},h_{1:N}^{t-1}) 
  &= \prod_{n=1}^N \Cat(a_n^t|\fCLS(h_n^t))
  \delta(h_n^t | \fRNN(\tilde{h}_n^{t-1},a_n^{t-1}))
  \delta(\tilde{h}_n^{t-1} | \fGNN(h_{1:N}^{t-1}))
\end{align*}
\caption{Formal specification of \modelshort\ model.
$h_n^t$ is the hidden state of RNN $n$ at time $t$;
$\delta(a|b)$ denotes a deterministic probability distribution.
Here $\fRPN$ is a region proposal network applied to frame $V^0$ which predicts
the location of $N$ boxes $b_{1:N}^0$.
$\fROI$ is a region of interest feature extractor applied to S3D
features derived from frames $V^{-H:0}$ at the locations
specified by the boxes.
$\fCLS$ is an MLP classifier with softmax output,
and $\Cat(a|p)$ is a categorical distribution over action labels $a$
with parameters $p$.
$\fGNN$ is a graph neural network where the input nodes are the old RNN hidden states,
$h_{1:N}^{t-1}$, and the output nodes are denoted $\tilde{h}_{1:N}^{t-1}$;
its definition is given in Table~\ref{tab:GNN}.
Finally, $\fRNN$ is a recurrent neural net update function
applied to the previous predicted label and GNN output.
}
\label{tab:model}
\end{table*}

\begin{table*}
\begin{align*}
\tilde{h}_{1:N} &= \fGNN(h_{1:N})
  =
  \prod_{i=1}^N \delta(\tilde{h}_i | \fnode([h_i, z_i]))
  \delta(z_i|\sum_j \alpha_{ij} h_j)
  \prod_{j=1}^N \delta(\alpha_{ij} | \softmax_j(\fattn(e_{i,1:N})) 
  \delta(e_{ij}|\fedge(h_i, h_j))
\end{align*}
\caption{Formal specification of the graph neural network.
$\fnode$ is an MLP that computes node states,
$\fedge$ is an MLP that computes edge states,
$\fattn$ is a self-attention network,
and $\softmax_j(l)$ is the $j$'th output of the softmax function with logits $l$.
}
\label{tab:GNN}
\end{table*}

\subsection{Modeling the edges}
\label{sec3:drn}
\label{sec:drn}

Our current model captures action dynamics for a set of independent actors. Many actions involve interactions among actors (\eg hugging and kissing). To capture such interactions, it is important to model the relations between different actors. Motivated by the recent success on relational reasoning, we combine  graph neural networks (GNN) with recurrent models. We treat each actor as a node and use the RNN's latent representation $\mathbf{h}_i^t$ as node feature.

We first consider a general graph network definition. For simplicity, we ignore the time step $t$, and denote $\mathbf{h}_i$ as the feature representation for node $i$. Let's also denote $\mathcal{N}_i$ as the neighbors of $i$ in the graph. We compute the output representation $\mathbf{\tilde{h}}_i$ from the input features of the other connected nodes:
\begin{equation}
\mathbf{e}_{ij} = \fedge(\mathbf{h}_i, \mathbf{h}_j)
\end{equation}
\begin{equation}
\mathbf{\tilde{h}}_i = \fnode(\{ \mathbf{e}_{ij} : j \in \mathcal{N}_i\})
\end{equation}
where $\mathbf{e}_{ij}$ is the derived features for edge $(i,j)$. Both $\fedge(\cdot)$ and $\fnode(\cdot)$ can be implemented with neural networks. Note that $\fnode(\cdot)$ is a function mapping a set to a vector. To make it permutation invariant, it is often implemented as
\begin{equation}
\label{eq:node}
\mathbf{\tilde{h}}_i = \fnode\left( \frac{1}{|\mathcal{N}_i|} \sum_{j \in \mathcal{N}_i} \mathbf{e}_{ij}\right) .
\end{equation}
\ie the output node feature is the average over all edge features connected to the node.

The overall graph-RNN update function can thus be expressed as
\begin{equation}
\mathbf{h}_i^t = \fRNN(\mathbf{\tilde{h}}_i^{t-1}, \mathbf{a}_i^{t-1})
\label{eq:rnn}
\end{equation}
\ie we apply GNN on the hidden states at time $t-1$, then run the standard RNN update function.

In practice, the number of actors are not known before hand, neither are the graph structures (relations) provided as supervision. Besides, the outputs from person detectors are typically over-complete and noisy (\eg hundreds to thousands of proposals per frame are generated by Faster-RCNN). One method to handle unknown graph is via ``relation networks''~\cite{RN_deepmind17,acrn}, which assumes the graph is fully connected (\ie $\mathcal{N}_i = \{\mathbf{v}_j \ |\ j\neq i\}$). According to Equation~\ref{eq:node}, this leads to an average over all edge features, and is sensitive to noisy nodes.

To mitigate this problem, we introduce the concept of ``virtual node'' $\mathbf{z}_i$ for node $i$. The virtual node is connected to all nodes in $\mathcal{N}_i$, and aggregates the node features with a weighted sum:
\begin{equation}
\mathbf{z}_i = \sum_{j} \alpha_{ij} \mathbf{h}_j
\end{equation}

The distributions of soft weights for neighboring nodes are given by
\begin{equation}
\alpha_{ij} = \text{softmax}(\fattn(\mathbf{e}_{ij}))
\end{equation}
where $\fattn(\cdot)$ is a attention function that measures the importance of node $j$ to node $i$. $\fattn(\cdot)$ can be efficiently implemented as the self-attention mechanism~\cite{Vaswani2017,Velickovic2018} with neural networks. Its parameters can be jointly learned with the target task using back propagation, and thus requires no additional supervision.

Once $\mathbf{z}_i$ is computed, we assume node $i$ is connected only to this virtual node, and updates the output feature by
\begin{equation}
\mathbf{\tilde{h}}_i = \fnode([\mathbf{h}_i; \mathbf{z}_i])
\label{eqn:hnode}
\end{equation}
The difference from graph attention networks
\cite{Velickovic2018}
is that they use $\mathbf{\tilde{h}}_i = \fnode(\mathbf{z}_i)$.
We have found this gives worse results (see Section~\ref{sec:results}),
perhaps because the model is not sure if it should focus on  features from itself or features from neighbors.
 
\subsection{Summary of model}

We call our overall model \model or \modelshort for short.
See Figure~\ref{fig:flow_chart} for a sketch,
and Tables~\ref{tab:model} and \ref{tab:GNN} for a precise specification of the model.

\subsection{Training} 
\label{sec3:overall}

The overall framework is jointly optimized end-to-end, where the loss function is
\begin{equation}
\mathcal{L}^\text{total} = \alpha\mathcal{L}^\text{loc} + \sum_{t=0}^T\beta_t\mathcal{L}^\text{cls}_t
\end{equation}

Here $\mathcal{L}^\text{loc}$ is the box localization loss given by the region proposal network and the bounding box refinement network computed for the last observed frame. $\mathcal{L}^\text{cls}_t$ is the action classification loss at time $t$. $\alpha$ and $\beta_t$ are scalars which balances the two sets of losses. In practice, one may want to downweight  $\beta_t$ for larger $t$ as the prediction task becomes more challenging.

Note that we do not use teacher forcing during training, to encourage the model to predict multiple steps into the future.
That is, when computing the predicted labels $a_i^t$, we condition on previous \emph{predicted} labels $a_{1:t}^{0:t-1}$
rather than the ground truth predicted labels.
(We use the soft logit scores for the predictions, to avoid the need to sample from the model during training.)

\begin{table*}[t!]
\begin{center}
\begin{tabular}{c|cc|cccccc}
\toprule
Method & Dynamics Model & Relation Model & $t=0$ & $t=1$ & $t=2$ & $t=3$ & $t=4$
& $t=5$ \\
\midrule
Single-head & - & - & 19.1 & 7.8 & 5.3 & 4.2 & 2.6 & 1.8\\ 
Multi-head & - & - & 16.0 & 9.4 & 6.8 & 5.4 & 4.3 & 3.6\\
GRU & GRU & - & 18.7 & 13.1 & 10.3 & 8.0 & 6.7 & 5.7\\
Graph-GRU & GRU & RN~\cite{RN_deepmind17} & 17.3 & 12.3 & 9.9 & 7.7 & 6.5 & 5.3\\
Graph-GRU & GRU & GAT~\cite{Velickovic2018} & 16.4 & 12.3 & 9.3 & 7.3 & 6.2 & 5.2\\
Graph-GRU & GRU & \modelshort (Us) & \textbf{20.4} & \textbf{14.4} & \textbf{11.2} & \textbf{9.3} & \textbf{7.5} & \textbf{6.8}\\
\bottomrule
\end{tabular}
\end{center}
\caption{Ablation study on the AVA dataset. We report mean AP@.5 for six different time steps. $t=0$ corresponds to the last observed frame, i.e., standard action detection. Each step is one second long.}
\label{tab:ava_ablation}
\end{table*}

\subsection{Implementation details}
\label{sec3:implementation}
\label{sec:details}

Our implementation of the Faster-RCNN~\cite{ren2015faster,huang2016coco} proposal extractor largely follows the design choices of~\cite{acrn}. The region proposal network (RPN) uses a 2D ResNet-50 network and takes a single image as input. It is jointly trained with the whole framework, using the human annotations from target dataset. We apply RPN on the final observed frame of each example to generate actor proposals. To handle temporal context, the feature network uses an S3D-G~\cite{s3dg_2017} backbone, which is a type of ``inflated'' 3D convolutional neural network~\cite{i3d_cvpr17}, and takes sequences of frames as inputs. We apply the feature network to frames at $-H:0$, where $0$ is the frame number of the last observed frame, and $H$ is the duration of temporal context. Once the features are extracted, we apply a temporal convolutional layer after the \texttt{Mixed\_4f} layer of S3D-G to aggregate the 3D feature maps into 2D and then apply the standard 2D ROIPooling to crop out the features inside actor proposals. Each cropped feature map is passed to the remaining layers of S3D-G. The final outputs are average-pooled into 1024-dim feature vectors. 

The weights of ResNet-50 used by proposal network are initialized with an ImageNet pre-trained model. We keep top 300 proposals per image. The weights of S3D-G used by feature network are pre-trained from Kinetics-400~\cite{kinetics17}. The weights of newly added layers are randomly initialized from truncated normal distributions. Unlike otherwise mentioned, the inputs to the feature network at 10 RGB frames resized to 400 by 400. We use gated recurrent units (GRU)~\cite{gru} as the particular RNN architecture to model action dynamics. We set the number of hidden units to 1024, which is the same dimension as the visual input features. For the discriminative relation network, we implement $\fedge(\cdot)$ and $\fattn(\cdot)$ as single fully-connected layers.

During both training and inference, the input actions $\mathbf{a}_i^{t}$ to the GRU are generated by the model rather than provided by the ground truth. We set the localization weight $\alpha=1$. The classification weight $\beta_0$ is set to $1$, we linearly anneal $\beta_t$ such that $\beta_t=0.5$. To compute classification loss, we use softmax cross entropy for J-HMDB and the sum of sigmoid cross entropies for AVA
(since the action labels of AVA are not mutually exclusive). We optimize the model with synchronous SGD and batch size of 4 per GPU, and disable batch norm updates during training. We use 10 GPUs in total. Two techniques are used to stabilize and speed up training: first, we warm-start the learning rate from 0.008 to 0.08 for $T_w$ steps and use cosine learning rate decay~\cite{cosine_lr} starting from 0.08 for another $T_c$ steps; second, we apply a gradient multiplier of 0.01 to gradients computed from \modelshort to the feature map. For AVA, we set $T_w$ to 5K and $T_c$ to 300K. For J-HMDB, we set $T_w$ to 1K and $T_c$ to 40K.

\begin{figure*}
    \centering
    \includegraphics[width=.9\linewidth]{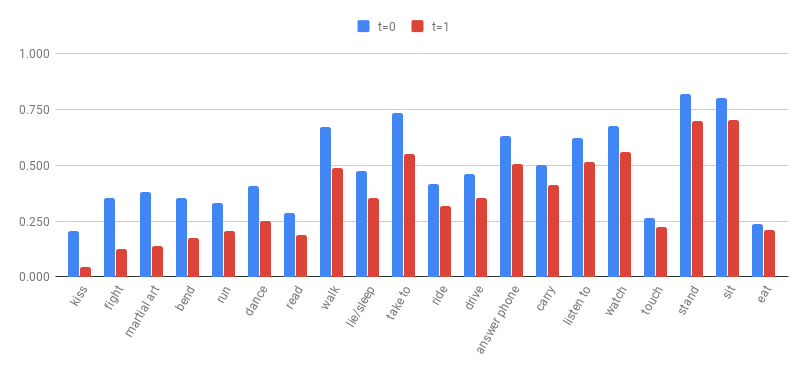}
    \caption{Change in AP performance  from $T=0$ to $t=1$.
    The actions with the biggest change are the hardest to predict in the future.}
    \label{fig:per_class_time}
\end{figure*}

\begin{figure*}
    \centering
    \includegraphics[width=.9\linewidth]{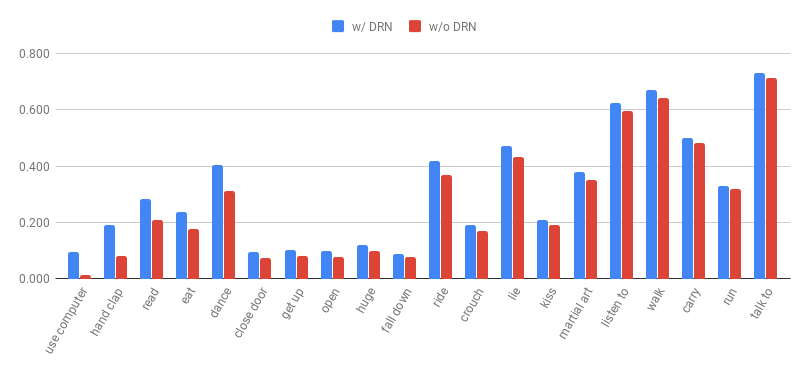}
    \caption{Change in AP performance  from adding graph connections  at $t=0$.
    The actions with the biggest change benefit the most from contextual modeling.}
    \label{fig:per_class_drn}
\end{figure*}

\begin{figure*}
    \centering
    \includegraphics[width=.9\linewidth]{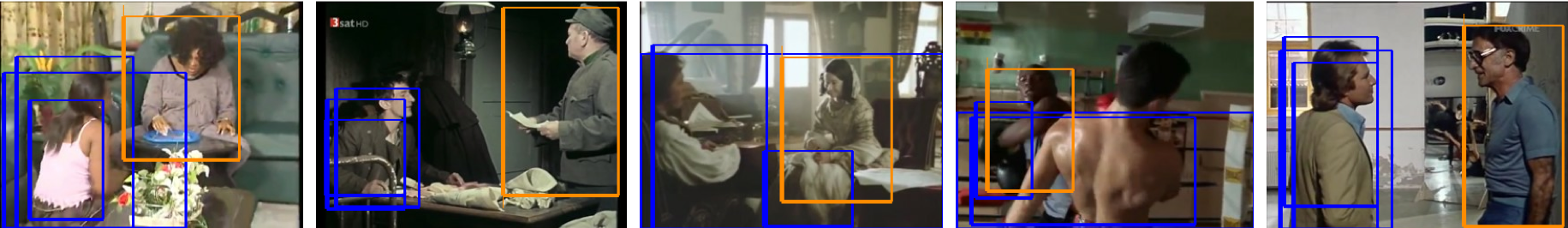}
    \caption{Visualizations of top 3 relations (blue boxes) selected for an actor proposal (orange box) by \modelshort on AVA. We see that the attended regions provide useful contextual information.}
    \label{fig:drn_viz}
\end{figure*}

\begin{figure*}
    \centering
    \includegraphics[width=.99\linewidth]{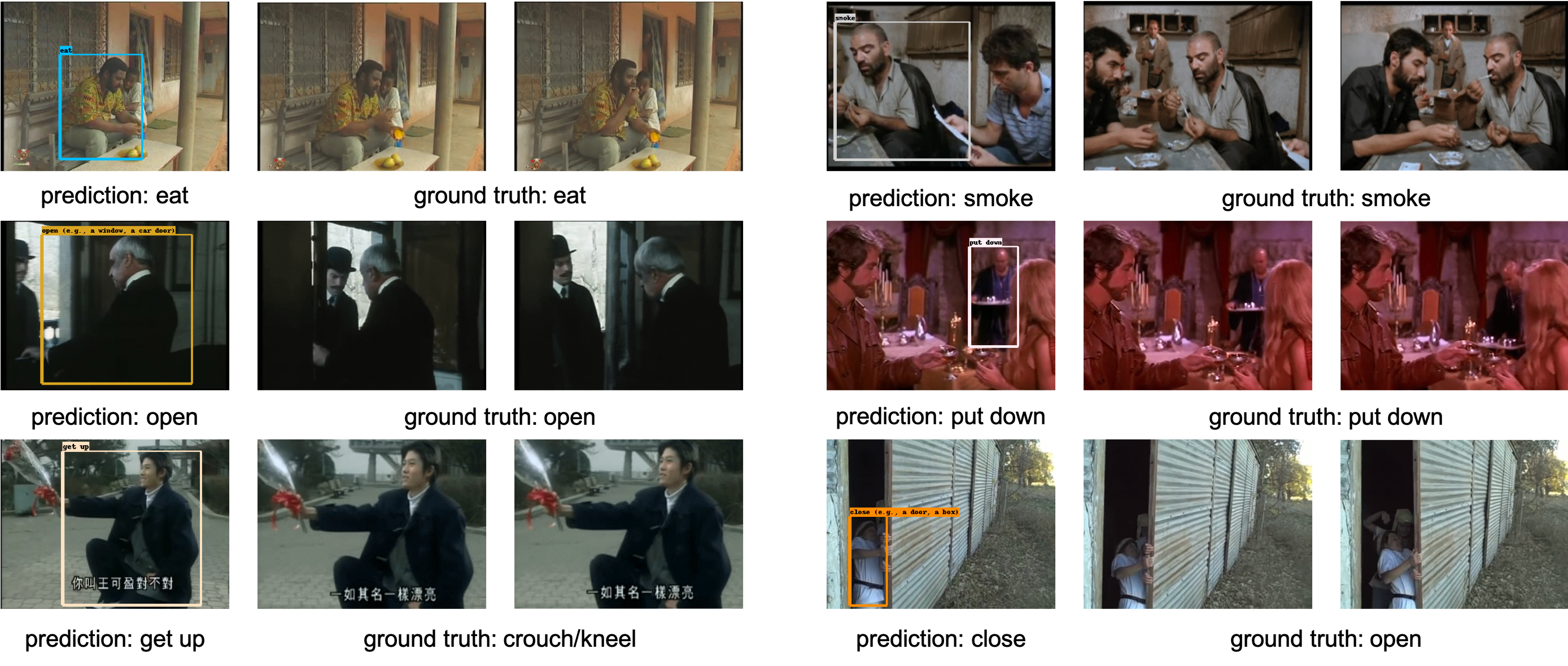}
    \caption{Example predictions on the AVA validation set at $t=1$. We show last observed frames at $t=0$ and render the detected actor boxes on the frames. To the right of each example, we also show the unobserved future frames half and one second ahead. We show top one detections if above threshold of 0.1, and remove the most frequent categories, such as sitting and standing. The top row shows examples where the model can predict what will happen in the future based on the current scene context. The second row shows examples where the model can predict what will happen in the future based on the other actors in the scene. The third row highlights the challenges in action forecasting (\eg multiple possible futures).}
    \label{fig:ava_final_viz}
\end{figure*}

\section{Experiments}
\label{sec:results}

In this section, we conduct experiments to study the impact of different design choices for relational reasoning and temporal dynamics modeling for the action prediction task.

\subsection{Experimental setup}

In the following we present the two datasets used in our experiments, 
Atomic Visual Actions (AVA)~\cite{ava_cvpr18} and 
J-HMDB~\cite{jhmdb} as well as the metrics used for evaluating action prediction.

{\bf AVA}~\cite{ava_cvpr18} is a recently released large-scale action detection dataset with 60 action classes. AVA is sparsely labeled at 1 FPS and each frame may contain multiple actors with multiple action labels. We use the most recent AVA version~2.1, which contains 210K training examples and 57K validation examples.  To get the ground-truth for future action labels, we use the bounding box identities (tracks) semi-automatic annotated for this dataset~\cite{ava_cvpr18}. 
For actor-level action prediction on AVA, we measure the IoU of all the actor detections with the ground-truth boxes. If the IoU is above 0.5, and the action label is correct, we consider it a true positive. For prediction, the ground-truth labels come from future boxes that are linked to the last observed ground truth boxes. We compute the average precision, and report per frame-level mean AP for different time steps.

{\bf J-HMDB} is an action detection dataset with 928 clips and 21 categories. Clips have an average length of 1.4 seconds.  Every frame in J-HMDB contains one  annotated actor with a single action label. There are three train/val splits for J-HMDB. We report results by averaging over the three splits, which is the standard practice on this dataset. Since there is only one action performed in each example, we follow the setup of Soomro \etal~\cite{Singh_ICCV2017,Soomro_online} and treat it as an early action prediction problem.
We report accuracy@K, which is the action classification accuracy by watching the first K\% of the videos.

\subsection{Action prediction on AVA}

This section presents and discusses quantitative and qualitative results for action prediction on the AVA dataset. We consider the following approaches:

\begin{itemize}
\setlength\itemsep{0em}
\item\textbf{Single-head}: for each future step $t$, train a separate model that directly classifies the future actions from visual features $\mathbf{v}_i$ derived from $V^0$.

\item\textbf{Multi-head}: similar to the single-head model, but jointly trains the classifiers for all $t$ in the same model. The visual features $\mathbf{v}_i$ are shared with all classification heads.

\item\textbf{GRU}: future actions are predicted from hidden states of GRU, where the states are initialized from visual features of the actor proposals. Model is trained jointly for all $t$.
However, there are no edges in the graph, so all nodes evolve independenty.

\item\textbf{Graph-GRU}: Same as GRU, but with a fully connected graph.
We consider 3 versions:
the Relation Network (RN)~\cite{RN_deepmind17}, which assigns equal weights to all pairwise relations;
Graph Attention Network~\cite{Velickovic2018}, which uses   a weighted sum of features from itself and all its neighbors;
and our proposed method, which uses Equation~\ref{eq:node}.

\end{itemize}

The results are shown in Table~\ref{tab:ava_ablation}. The first three rows compare the impact of different approaches for dynamics modeling. Our first observation is that, as $t$ grows larger, the mean AP declines accordingly. This is expected since the further away in time the prediction is, the harder is the task.
Our second observation is that the single-head baseline performs worse on all $t$ except for $t=0$, where the frames are observed. The lower performance of $t=0$ for multi-head can be explained by the fact that the joint model has less model capacity compared with 6 independent single-head models. However, we can see that by sharing the visual features in a multi-task setting, the multi-head baseline outperforms its single-head counterpart for future prediction at $t > 0$.
Our third observation is that using GRU to model action dynamics offers better future prediction performance without sacrificing detection performance at $t=0$,
since it can capture patterns in the sequence of action labels.

The last three rows of Table~\ref{tab:ava_ablation}
compares the impact of different relational models. We can see  DRN outperforms the other two consistently. For RN, one possible explanation for the performance gap is that it assigns equal weights to all edges, which is prone to noise in our case, 
since many nodes correspond to background detections.
For GAT, we notice that the performance at $t=0$ is much lower, indicating that it has difficulty 
distinguishing node features from neigbhor features.

Figure~\ref{fig:per_class_time} compares the classes with biggest performance drops from detection ($t=0$) to prediction ($t=1$).
We see that it is challenging for the model to capture  actions with short durations, such as kissing or fighting. Actions with longer durations, sich as talking, are typically easier to predict.
Figure~\ref{fig:per_class_drn} compares the effectiveness of \modelshort over the GRU baseline, without any edges in the graph. We can see that the categories with the most gains are those with explicit interactions (\eg hand clap, dance, martial art), or where other actors provide useful context (\eg eat and ride).
In Figure~\ref{fig:drn_viz}, we show the top 3 boxes ( blue) with the highest attention weights to the actor being classified (orange). We can see that they typically correspond to other actors. Finally, we visualize example predictions in Figure~\ref{fig:ava_final_viz}.

\subsection{Early action prediction on J-HMDB}

Finally, we demonstrate the effectiveness of \modelshort on the early clip
classification.  During training, we feed 10 RGB frames to the feature network, and predict one step into the future. During inference, we feed the first $K\%$ frames to the feature network, and take the most confident prediction as the label of the clip. Table~\ref{tab:jhmdb_ablation} shows the results, we can see that our approach significantly outperforms previous state-of-the-art methods. To study the impact of relation models, we also compare with the GRU only and the GAT baselines, and find \modelshort outperforms both. By inspecting the edge attentions, we observe that some of the RPN proposals cover objects in the scene, which are utilized by \modelshort to model human-object relations.

\begin{table}
\small
\begin{center}
\begin{tabular}{c|cccccc}
\toprule
Model & 10\% & 20\% & 30\% & 40\% & 50\% \\
\midrule
Soomro \etal~\cite{Soomro_online} & $\approx 5$ & $\approx 12$ & $\approx 21$ & $\approx 25$ & $\approx 30$ \\
Singh \etal~\cite{Singh_ICCV2017} & $\approx 48$ & $\approx 59$ & $\approx 62$ & $\approx 66$ & $\approx 66$ \\
GRU & 52.5 & 56.2 & 61.1 & 65.2 & 65.9 \\
GAT~\cite{Velickovic2018} & 58.1 & 61.8 & 64.4 & 68.7 & 68.8 \\
\modelshort & \textbf{60.6} & \textbf{65.8} & \textbf{68.1} & \textbf{71.4} & \textbf{71.8} \\
\bottomrule
\end{tabular}
\end{center}
\vspace{-.2in}
\caption{Early action prediction performance on J-HMDB.}
\label{tab:jhmdb_ablation}
\end{table}
\section{Conclusion}

We address the multi-person action forecasting task in videos. We propose a model that jointly models temporal and spatial interactions among different actors with \model. Quantitative and qualitative evaluations on AVA and J-HMDB datasets demonstrate the effectiveness of our proposed method.

\newpage

{\small
\bibliographystyle{ieee}
\bibliography{cvpr}
}

\end{document}